\pdfoutput=1
\documentclass[11pt]{article}

\usepackage{acl}

\usepackage{times}
\usepackage{latexsym}

\usepackage[T1]{fontenc}
\usepackage[utf8]{inputenc}

\usepackage{microtype}

\usepackage{graphicx}
\usepackage{subcaption}
\usepackage{mwe}
\usepackage{xcolor}
\usepackage{booktabs}
\usepackage{ifluatex,ifxetex}
\ifluatex
    \usepackage{fontspec,luatex85}
\else
   \ifxetex
       \usepackage{fontspec}
   \else
       \usepackage[T1]{fontenc}
   \fi
\fi

\usepackage{enumitem}
\usepackage{titlesec}

\title{{R}icher {C}ountries and {R}icher {R}epresentations}

\author{Kaitlyn Zhou \\
    Stanford University \\
    katezhou@stanford.edu \\\And
  Kawin Ethayarajh \\
  Stanford University\\
   kawin@stanford.edu \\\And
  Dan Jurafsky \\
  Stanford University\\
  jurafsky@stanford.edu}

\begin{document}
\maketitle
\begin{abstract}
% How does BERT represent entities and what factors influence its representations? Given known imbalances in training datasets, how do low and high frequency entities vary in their geometric representations and what are the potential downstream consequences? 

% The literature on bias in language representations has largely focused on finding undesirable associations (e.g., between gender pronouns and gender stereotypes). 
% In this work, we focus on country names, an information-rich entity,  we present another axis along which bias can manifest: the diversity of representations. 
% We find that only a subset of country names are in BERT’s vocabulary and that tokenization greatly affects the representation of low frequency country names. 
% Training word frequency explains this effect but tokenization amplifies real world data imbalances. 
% As a consequence, we find that the volume occupied by their embeddings correlates with frequency and thus GDP, with richer countries having more varied representations. 
% Additionally, we find that poorer countries are seen as less distinguishable and their existence less recognized e.g., Ghana (the correct answer) is not in the top five predictions for, ``The country producing the most cocoa is [MASK].’’. 
% We experiment Multilingual-BERT and fine-tuning as potential mitigation strategies but find limitations to their effectiveness. Finally, we recommend increasing transparency of training word frequencies and recommend the community to consider defining and designing \textit{representational guarantees}.

We examine whether some countries are more richly represented in embedding space than others. We find that countries whose names occur with low frequency in training corpora are more likely to be tokenized into subwords, are less semantically distinct in embedding space, and are less likely to be correctly predicted: e.g., Ghana (the correct answer and in-vocabulary) is not predicted for, ``The country producing the most cocoa is [MASK].''. Although these performance discrepancies and representational harms  are due to frequency, we find that frequency is highly correlated with a country's GDP; thus perpetuating historic power and wealth inequalities. We analyze the effectiveness of mitigation strategies; recommend that researchers  report training word frequencies; and recommend future work for the community to define and design \textit{representational guarantees}.

% Cosine similarity of BERT embeddings are used in a variety of NLP tasks (e.g., QA, IR, MT) and metrics (e.g., BERTScore). However it isn’t clear if these similarity scores are well calibrated to human perception, or whether there are systematic ways in which some word relationships are exaggerated or understated. We investigate this by proposing two novel tools to explore the similarity and geometric characteristics of contextualized word embeddings: (1) an \textit{identity probe} that predicts the identity of a word using its embedding and (2) the minimal bounding sphere for a word's various contextualized representations. The first assesses how lexical properties affect a word’s embedding and the second measures the geometric space occupied by such embeddings. Together, our results reveal that words of high and low frequency have significant differences in their representational geometry. Such differences introduce distortions: when compared to human judgments, point estimates of embedding similarity (e.g., cosine similarity) can over- or under-estimate the semantic similarity of two words. Most importantly, these distortions vary with respect to the frequency of those words in the training data. This has downstream societal implications: BERT-Base has more trouble differentiating between South American and African countries than North American and European ones. We find that these distortions persist when using BERT-Multilingual, suggesting that they cannot be easily fixed with additional data, which instead introduces new distortions.
\end{abstract}

\section{Introduction}
How similar are the words ``Brooklyn'' and ``Queens''? To a New Yorker, they evoke two very different places, cultures, and cuisines, but to a Seattleite, they are quite similar, both being boroughs of New York City.\footnote{\url{https://en.wikipedia.org/wiki/View_of_the_World_from_9th_Avenue}.} Our perception of  entities such as cities or countries is conditioned on our backgrounds. Here, we ask if language models are also susceptible to representational biases.

We suggest three criteria to characterize the quality of representations for particular entities or groups: consistency, distinctiveness, and recognizability. For consistency, are all entities of a certain type (such as all country names) represented with the same number of tokens in the lexicon? For distinctiveness, are entities of the same category seen as equally distinct in representational space? For recognizability, are models capable of generating all entities of a certain type in response to questions?  And are the differences between entities confounded across lines of historical inequity (like wealth of countries)?

Focusing on BERT (\texttt{bert-base-cased}\footnote{\url{https://huggingface.co/bert-base-cased}}) representations \cite{devlin-etal-2019-bert}, we find that names of countries that appear less frequently in training data are less likely to be in-vocabulary, are less semantically distinct from other countries, and are less frequently predicted in the masked language modeling (MLM) task. Disappointingly, we find similar behavior in  \texttt{bert-base-multilingual-cased} and \texttt{roberta-base}. We identify these differences as intrinsic representational harms where low frequency countries are more likely to be conflated with one another and their existence less recognized. 

A more troubling result is that training data frequency is highly correlated with the gross domestic product of a country (GDP) (Pearson’s $r = 0.82$). Our training data and thus the representation of entities through our language models encodes wealth and power disparities and perpetuates representational harms.
Given these significant differences in representation, what could it look like to impose a minimum quality of representation for significant entities? We recommend that the community consider designing \textit{representational guarantees} for language models. 

In summary we: 1) reveal multiple ways in which the BERT representation of high  GDP countries is systematically richer than that of low GDP countries; 2) study the effectiveness of potential mitigation efforts; and 3) propose the idea of \textit{representational guarantees} as future work for the community.

\section{Related Work}

Language technologies have long been studied for potential intrinsic and extrinsic harms \citep{galliers1993evaluating}. 
Known intrinsic harms include misrepresentation of gender \citep{NIPS2016_a486cd07}, race \cite{abid2021persistent}, and ability \citep{hutchinson-etal-2020-social} --- all types of representational harms  \citep{barocas2017, crawford2017trouble, blodgett-etal-2020-language}. Other intrinstic harms include forms of erasure through under-representation of LGBTQ+ identity terms \cite{queering, oliva2021fighting} and racial groups \cite{gehman-etal-2020-realtoxicityprompts}. Extrinsic harms are often found in downstream tasks and include disparities in quality of service among user groups \cite{10.1145/3368555.3384448} such as African-American users \cite{blodgett2017racial, Koenecke2020RacialDI}. However, low statistical power has also made it difficult to make conclusive claims about the presence or absence of bias \citep{ethayarajh-2020-classifier}.

Many of these representational harms have been linked to word frequency in static embeddings \cite{NIPS2016_a486cd07, Caliskan183, zhao-etal-2018-gender, bordia-bowman-2019-identifying,  ethayarajh-etal-2019-understanding, van2022regional}. Low frequency words also differ geometrically from other words, with  smaller inner products \citep{mimno-thompson-2017-strange} and lower variance \citep{ethayarajh-etal-2019-towards}. Recent work has also shown how frequency impacts \textit{contextual} embeddings such as the under-estimation of cosine similarity among high-frequency words \citep{zhou_cosine_etal_2022} and the discrepancies in representations of personal names \cite{shwartz-etal-2020-grounded, wolfe-caliskan-2021-low}. Our work extends these lines of work via the examination of representational harms for country names.

% We build on past works of intrinsic harms of language technologies and examine the effects of topic skews in training data corpus and how it affects the representation of country names.
 
% The impact of frequency has long been studied on static word embeddings \citep{NIPS2014_feab05aa, hellrich-hahn-2016-bad, mimno-thompson-2017-strange, wendlandt-etal-2018-factors}. For example, 

% \citet{bandy2021addressing} found significant skews in genre representation in BookCorpus and \citet{whatsinwiki} found that culture, arts, and people to be the most common categories of Wikipedia. These two corpora make up the training data in BERT-base and their topic distributions likely has significant impacts on what is learned in our language models. 
% \section{Methods and Data}
% We used the March 1, 2020 Wikimedia Download and BookCorpus frequencies \citep{7410368, hartmann-dos-santos-2018-nilc} to approximate word frequencies in the BERT pre-training corpus. We refer to this as a word's \textit{training data frequency}.

% We use sentences from English Wikipedia to extract word embeddings by averaging the last four hidden layers of BERT. In the case where words were split into multiple parts from the tokenizer, we take the average embedding of the subwords. \footnote{We matched keywords on token IDs to ensure punctuation and casing are consistent across examples.}
\section{Rich Countries have their own Tokens}
\label{section:token}
Are poor and rich countries tokenized the same way? Here, we focus on BERT's tokenization process and measure the consistency (or rather inconsistency) in how names of countries are tokenized and then represented. Our GDP data is retrieved from the  United Nations Statistics Division from 2019 \footnote{\url{https://unstats.un.org/unsd/snaama/Basic};
GDP data was shown in USD.}\footnote{Code for this paper can be found at \url{https://github.com/katezhou/country\_distortions}}.

Of the 159 single-word countries names from the United Nations members list, 134 of them are in-vocabulary, ---  the remaining 25 are out-of-vocabulary (OOV). In  WordPiece tokenization,  OOV words are tokenized into in-vocabulary subwords (e.g. ``Andorra'' becomes ``And'' and  ``\#\#orra'', see table \ref{table:tokens}). Additionally, as a limitation of the unigram vocabulary, the 34 multi-word country names (e.g. ``United States") are also OOV and represented as distinct tokens (``United'' and ``States''.)\footnote{Acronyms of some high GDP countries e.g. US, USA, UK, UAE are in-vocabulary but are not included in this study to avoid introducing confounders concerning which to include and how to pool. This is likely a conservative bias since including them would have increased the effects we study.} Each word of multi-word countries can also be OOV (e.g., Sao Tome and Principe is tokenized into 9 different subwords).

We used ordinary least squares regression to predict the number of subword tokens in each country name, using training data frequency of each country name as the feature (BERT training data estimated from the March 1st, 2020 Wikimedia Download and BookCorpus) \citep{7410368, hartmann-dos-santos-2018-nilc}.\footnote{Additional tools used: \url{https://github.com/IlyaSemenov/wikipedia-word-frequency}; 
\url{https://github.com/attardi/wikiextractor}}.  We found that training data frequency explains 38\% of the variance in number of subwords ($p < 0.01$), despite the confounder of multi-word countries being considered OOV (Table \ref{table:norm_regression} in Appendix). This is likely due to the fact that the tokenizer builds its vocabulary based on fitting training likelihood. Given that the training data frequency of a country's name correlates strongly  with its GDP ( Pearson's $r=0.82$), the countries that have the highest number of subwords are also the ones with the lowest GDP.

\begin{table}[]
\begin{center}
\begin{tabular}{cccc}
\midrule
\textbf{Subwords} & 
\textbf{Freq} & 
\textbf{GDP (M)} & 
\textbf{Example} \\
\midrule
1 (n=134) & 74,882 & 430,596 & Uzbekistan \\
2 (n=32)& 68,148 & 870,702 & Comoros \\
3 (n=15) & 8,711 & 34,896 & Grenada \\
4+ (n=12) & 4,309 & 14,980 & Eswatini\\
\midrule
\end{tabular}
\vspace{-2mm}
\caption{The average BERT training data frequency and GDP associated with the countries, binned by the number of subwords the country was tokenized into (e.g., ``Grenada'' was tokenized into three subwords.). Using OLS to predict number of subwords, frequency explains 38\% of the variation in number of subwords.}
\label{table:tokens}
\end{center}
\vspace{-5mm}
\end{table}

\begin{figure}[htb]
    \centering
    \vspace{-3mm}
     \includegraphics[width=.45\textwidth]{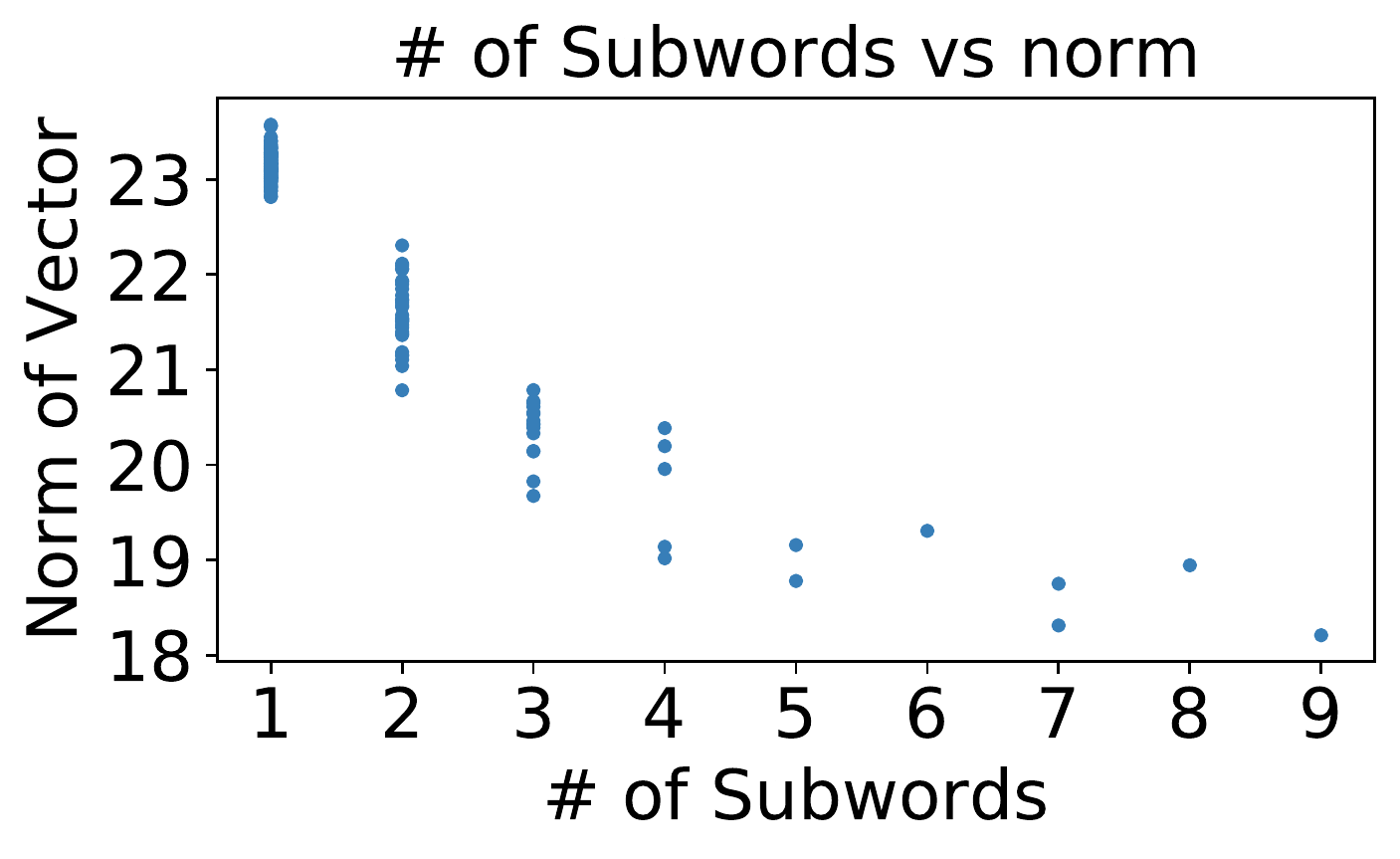}
     \vspace{-3mm}
     \caption{Average norm of embeddings in relation to the number of subwords of the embedding. OOV words are represented by averaging subwords. Pearson's correlation between average norm and number of subwords, $r=-0.92$.}
     \label{figure:norm_freq}
     \vspace{-2mm}
\end{figure}

\subsection{Geometric Impact of being OOV}
The methods used to represent OOV words can  impact the geometry of their representations. For example, the representation of OOV words often have smaller norms ($L^2$ of the word embedding) than in-vocabulary words. This is because a common way to represent OOV words is to take the average of their subwords \citep{pilehvar-camacho-collados-2019-wic, blevins-zettlemoyer-2020-moving, bommasani-etal-2020-interpreting}. However, since the average values for each dimension are near zero, the more subwords that are averaged,  the smaller the norm of the vectors \cite{adi2016fine} (figure \ref{figure:norm_freq}). As a result, the Pearson's correlation between the norm and number of subwords is $r=-0.92$. A common alternative to averaging subwords is to use the first subword to represent an OOV word. When this representation is used, the correlation between the norm and the number of subwords reverses and is slightly positive, Pearson's $r=0.22$ (figure \ref{figure:norm_freq_first} in Appendix). One possible explanation could be that the first subword of OOV words are more likely to be stop words, which are known to have larger norms \citep{ethayarajh-2019-contextual}. The difference in geometry between OOV and in-vocabulary words exists in both representation methods. This could could result in impacts on tasks using embedding-based retrieval (e.g., nearest-neighbor LM) as low-frequency names will have additionally distinguishing geometric characteristics as a result of tokenization.

Inconsistency in tokenizing country names leads to inconsistency in geometric representation. A potential mitigation might be to have all country names be in-vocabulary with a dedicated token. This might not address all impacts of training data frequency imbalances, but would at least  prevent additional geometric differences due to tokenization.

\section{Richer Countries are Most Distinct in Embedding Space}
\label{section:distinct}
Nations or entire regions are subject to bias. The African continent, for example, is often treated journalistically as a single homogeneous entity \citep{Nothias_2018}, as if African countries are all substitutable for one another.   
We draw on this finding to ask whether historically disadvantaged countries are also conflated with one another in the embeddings of their names (i.e., seen as less distinct from each other than other countries).

We measure the semantic similarity between pairs of the 134 in-vocabulary countries by creating word embeddings for each name (done by averaging the last four hidden layers of BERT). We calculate the average in-group cosine similarity of country names as grouped by frequency (i.e., we take the countries in each decile of frequency and measure the average cosine similarity across the ${14 \choose 2}$ pairs). We repeat this ten times and find that the 10\% least frequent names have an average in-group similarity of 0.610 compared to an average similarity of 0.582 for the 10\% most frequent countries (mean $\delta=0.028$, permutation test $p<0.01$). 

\begin{figure}[t]
    \centering
    \vspace{-2mm}
     \includegraphics[width=.45\textwidth]{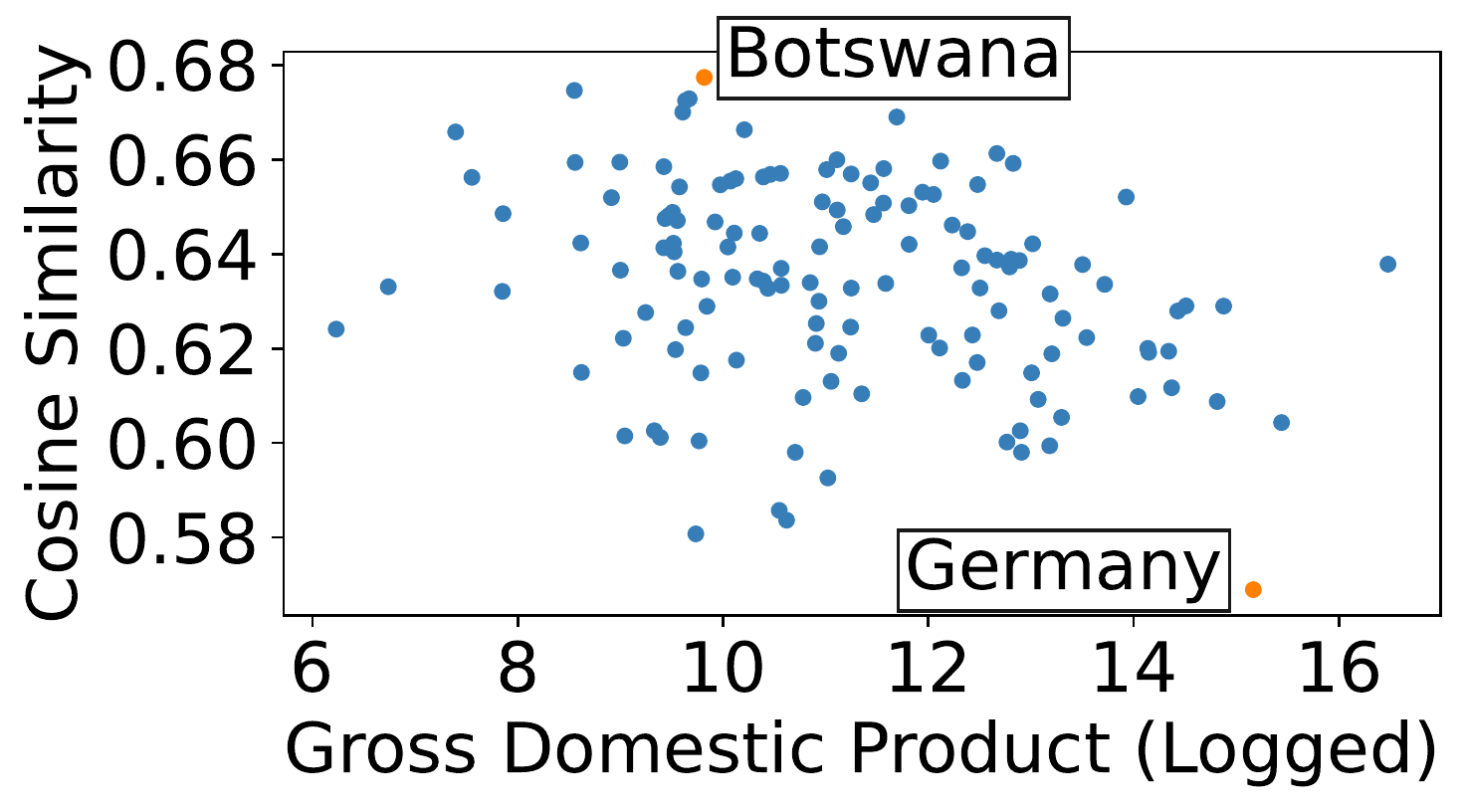}
     \vspace{-3mm}
     \caption{Average cosine similarity of a country to all other countries vs.\ its GDP. Pearson's $r=-0.28$ for average cosine similarity and GDP. Wealthier countries are more distinct in BERT embedding space.}
     \label{figure:gdp_cosine}
     %\vspace{-1 em}
\end{figure}

We then calculate the average  semantic similarity of a country's embedding to all other countries to measure a country's \textit{distinctiveness} (averaged over ten trials). Using OLS to predict average cosine similarity, frequency explains 8\% of the variance (Table \ref{table:freq_log_in_vocabulary} in Appendix). For example, in our experiments, France had a cosine similarity of $\geq0.7$ with 21 other countries while Haiti shared a cosine similarity of $\geq0.7$ with 59 other countries. France's distinctiveness contrasts with Haiti's similarity with other countries. Using these embeddings and cosine similarities in a downstream task like IR (or MT, where embedding cosines are used in algorithms like BERTScore) would yield vastly different results despite using the same threshold. We visualize how average cosine similarity correlates with a country's GDP in Figure \ref{figure:gdp_cosine}.

%  For contrast, when using static embeddings\footnote{\texttt{word2vec-google-news-300}} to compare the semantic similarity between countries, countries tend to not share high similarity with other countries. The Zambia is found to be the country that is most similar to other countries, sharing a cosine similarity with 9 other countries.

To ensure that the semantic similarity discrepancies are not simply a consequence of how these countries are written about in test examples, we run the same OLS experiment on an artificial dataset where names of countries appear in identical contexts. The results are consistent: frequency explains 9\% of the variance in average cosine similarity (Table \ref{table:freq_log_in_vocabulary_artificial} and Table \ref{table:artificial_sentences} in Appendix). Independent of \textit{how} these countries are being written about in any potential downstream task, NLP models like BERT embed country names in a way that results in higher semantic similarity for low frequency names, and hence low GDP countries. This is a representational harm:  distinctiveness of a nation's name in NLP representations correlates with the nation's wealth, resulting in poorer countries being more likely to be conflated with one another.

\subsection{Tokenization and Semantic Similarity} As discussed in the section above, tokenization results in geometric differences between OOV and in-vocabularly words, and here we examine tokenization's effect on cosine similarity. OOV country names have higher in-group similarity averages when averaging subpieces (0.668 vs 0.628; mean $\delta=0.040$, permutation test $p< 0.01$). 
Given that OOV words have smaller norms and are closer to the centroid, this signals a concentration of low-frequency words. However, when using the first subpiece to represent OOV words, OOV country names have lower in-group similarity averages than in-group similarity among in-vocabulary words (0.589 vs 0.625; mean $\delta=0.037$, permutation test $p< 0.001$) --- we showed that these embeddings conversely have larger norms and could be more widely dispersed. The key takeaway here is that both methods show semantic differences between in-vocabulary and OOV words; again there are potential impacts in embedding-based downstream tasks.

\section{Richer Countries are more Frequently Predicted}
\label{section:mlm}
The lack of recognition of people and groups \cite{queering, oliva2021fighting, gehman-etal-2020-realtoxicityprompts} has often been cited as an representational harm. Here, we use the masked language modeling task as a proxy for many downstream tasks that need to be able to predict the name of a country (e.g., as the answer to a question, or in a summary) to measure whether countries are all minimally predicted or whether instead we see representational harms such as erasure.

We randomly sample sentences from Wikipedia that contain the name of a country, replace the name with a BERT mask token (\texttt{[MASK]}), and use the masked-out country name as the gold label. We use 100 examples of each of the 134 in-vocabulary country names.\footnote{This is a significantly harder task than ordinary MLM as only names of countries are masked out, there could be multiple appropriate answers, and a limited context is given.} The model will only predict in-vocabulary words, which again illustrates the impact of inconsistent tokenization: the very task BERT was trained on is unable to handle OOV country names without modification.

\begin{figure}[t]
    \centering
    \vspace{-3mm}
     \includegraphics[width=.45\textwidth]{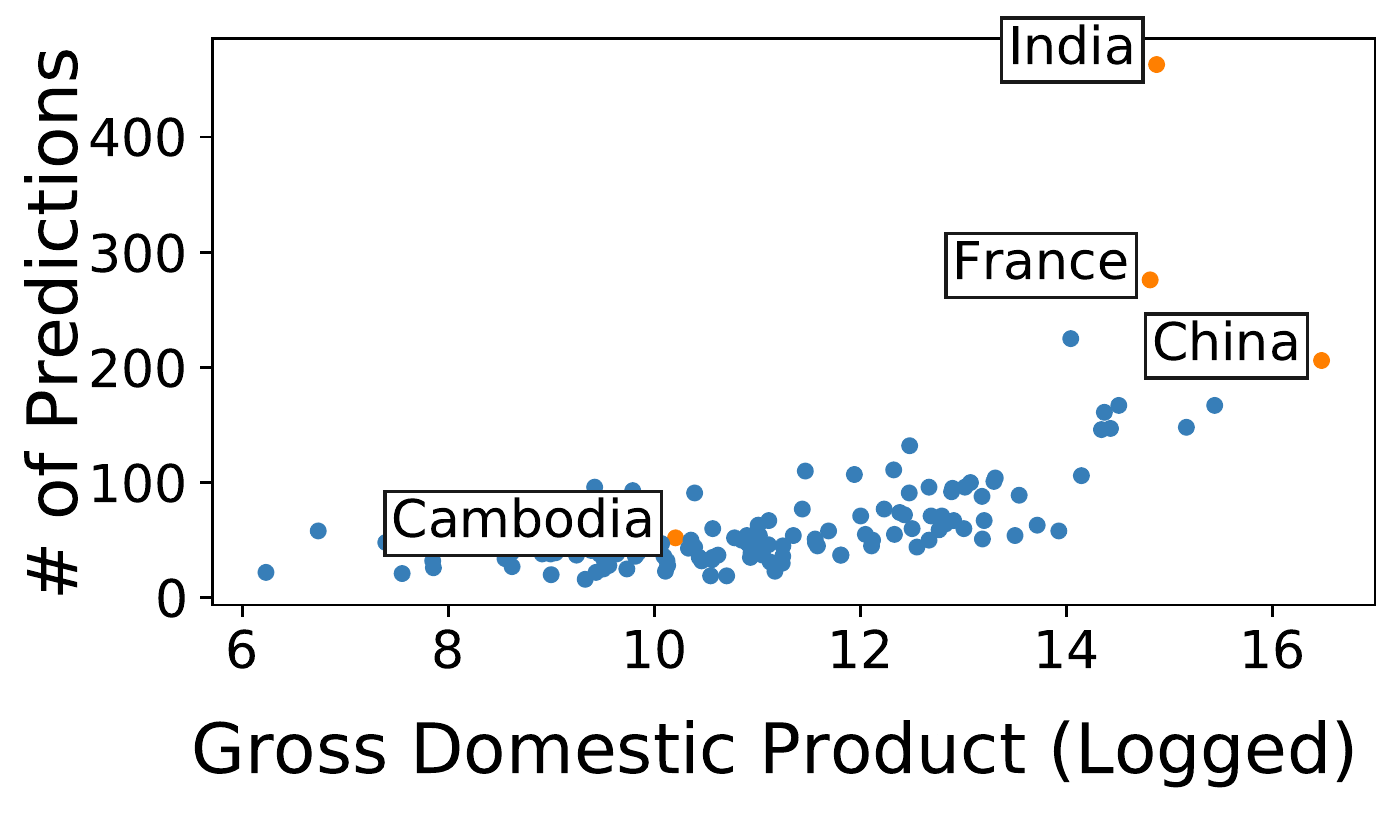}
     \vspace{-3mm}
     \caption{Number of times a country was predicted in the MLM task versus its GDP (logged). Pearson's correlation between GDP (logged) and number of times a country is predicted, $R=0.64$.}
     \label{figure:mlm_freq}
     \vspace{-2mm}
\end{figure}

% BERT performs best at predicting high frequency (high GDP) countries in the masked language modeling task.

High frequency countries (75th percentile) have an average accuracy of 42\% while low frequency (25th percentile) countries have an accuracy of 26\% (table \ref{table:performance_by_model} in Appendix). How often a country is predicted is highly correlated with its training data frequency (Pearson's $r=0.64$, figure \ref{figure:mlm_freq}). Of all the country names predicted, the 10 least predicted countries make up 2\% of total guesses compared to 25\% for the top 10 most predicted countries. BERT fails this task in drastic ways; predicting China, India, Brazil for, ``The poorest country in the world is [MASK].'' (Burundi and Somalia ranked as the poorest by GDP/capita). This reduced  recognition of poor countries is a representational harm.

% China, India, and Brazil have the 2nd, 6th, and 13th largest GDPs and are respectively 100th, 155th, and 115th richest by GDP/capita

%  predicting China, Bolivia, Brazil for, ``The country producing the most copper is [MASK].''; predicting: India, Brazil, France for ``The longest river in the world is in [MASK].''; and 
% The correct answers are: Chile, Egypt and Sudan, and Brunudi. 

% and instead predicts high GDP countries instead. Examples include: .

\section{The Limitations of Mitigation Efforts}
\label{section:mitigation}
We analyze the effectiveness of two popular mitigation efforts.

1. Could these frequency-based effects be mitigated with {\bf additional training data}?
We measure the performance of Multilingual BERT (BERT-ML) which includes Wikipedia articles from 104 other languages.  We continue to find accuracy disparities between low and high GDP countries (15\% vs 29\%). Additional training data fails to mitigate these harms, most likely because the additional data continues to amplify existing imbalances (i.e., German, French, and Polish are the next three biggest languages of Wikipedia articles). 
Data augmentation as a mitigation technique is challenging as datasets could easily maintain existing or introduce new frequency imbalances. We also tested RoBERTa (trained on over 160GB more  data) on this task with similar results (table \ref{table:performance_by_model} in Appendix).

2. Could fine-tuning or {\bf continuing pre-training} mitigate these harms?
We select 20 random countries (Appendix table \ref{table:list_of_interested_countries}); for each, we select 1,000 random sentences from Wikipedia that mentions the country's name. After four epochs of training, we then test on an evaluation dataset (100 examples/country) and we see a 13\% increase in performance on our selected countries and a 3\% decrease for other countries (table \ref{table:tuned_diff} and figure \ref{figure:fig_tuned} in Appendix). Our subset of interested countries originally made up 17\% of all guesses, but this rate more than doubles in our tuned model. This pre-training mitigation method shows promise but has trade-offs in performance, requires practitioners to be aware of inequalities, and have access to enough training samples to continue pre-training effectively.

\section{Discussion and Conclusion}
We find significant disparities in the quality of representation of country names and show how these differences result in representational harms that perpetuate existing wealth and power inequalities. We make two recommendations on paths forward.
We recommend the release of training word frequencies to increase transparency and isolate current representational harms \cite{gebru2021datasheets, 10.1145/3287560.3287596, bender-friedman-2018-data, ethayarajh2020utility}. Practitioners who use these models in their systems and research should have access to the topic and entity distribution of our models given the potential for frequency-related harms.
We also recommend the community consider designing \textit{representational guarantees} for significant entities to mitigate these downstream harms. Our work illustrates the potential harms that arise when entities such as country names do not have representational guarantees. We encourage the community to consider the following questions:  How can we ensure entities such as names of country be in-vocabulary? How can we guarantee a minimum distinctness and ensure recognition of historically disadvantaged groups? Such guarantees will likely be difficult to design and will require expertise from multiple domains but mitigating these representational harms is an important task that we cannot ignore. 

\section*{Acknowledgements}

We sincerely thank Justine Zhang, Rishi Bommasani, Tol\'{u}l\d{o}p\d{\'{e}} \`{O}g\'{u}nr\d{\`{e}}m\'{i}, and our anonymous reviewers for their support and helpful reviews. This research has been supported in part by a Hoffman-Yee Research Grant from the Stanford Institute for Human-Centered AI, award IIS-2128145 from the National Science Foundation, a Stanford Graduate Fellowship, a Facebook Fellowship, and the Natural Sciences and Engineering Research Council of Canada.
% Entries for the entire Anthology, followed by custom entries
\bibliography{custom}
\bibliographystyle{acl_natbib}

\appendix

\section{Appendix}
\label{sec:appendix}

\subsection{Appendix for section \ref{section:token}}

\begin{table*}
\begin{center}
\begin{tabular}{lclc}
\toprule
\textbf{Dep. Variable:}    &      \# of subwords      & \textbf{  R-squared:         } &     0.384   \\
\textbf{Model:}            &       OLS        & \textbf{  Adj. R-squared:    } &     0.380   \\
\textbf{Method:}           &  Least Squares   & \textbf{  F-statistic:       } &     118.8   \\
\textbf{Date:}             & Mon, 15 Nov 2021 & \textbf{  Prob (F-statistic):} &  7.95e-22   \\
\textbf{Time:}             &     16:41:27     & \textbf{  Log-Likelihood:    } &   -272.30   \\
\textbf{No. Observations:} &         193      & \textbf{  AIC:               } &     548.6   \\
\textbf{Df Residuals:}     &         191      & \textbf{  BIC:               } &     555.1   \\
\textbf{Df Model:}         &           1      & \textbf{                     } &             \\
\bottomrule
\end{tabular}
\begin{tabular}{lcccccc}
                      & \textbf{coef} & \textbf{std err} & \textbf{t} & \textbf{P$> |$t$|$} & \textbf{[0.025} & \textbf{0.975]}  \\
\midrule
\textbf{Constant}     &       7.1702  &        0.515     &    13.912  &         0.000        &        6.154    &        8.187     \\
\textbf{freq\_logged} &      -0.5502  &        0.050     &   -10.901  &         0.000        &       -0.650    &       -0.451     \\
\bottomrule
\end{tabular}
\begin{tabular}{lclc}
\textbf{Omnibus:}       & 90.415 & \textbf{  Durbin-Watson:     } &    2.177  \\
\textbf{Prob(Omnibus):} &  0.000 & \textbf{  Jarque-Bera (JB):  } &  339.064  \\
\textbf{Skew:}          &  1.900 & \textbf{  Prob(JB):          } & 2.36e-74  \\
\textbf{Kurtosis:}      &  8.265 & \textbf{  Cond. No.          } &     74.0  \\
\bottomrule
\end{tabular}
\caption{OLS Regression Results: Using training data frequency (logged) to predict number of subwords as tokenized by BERT. Training word frequency explains 38\% of the variance in number of subpieces as tokenized by BERT.}
\label{table:norm_regression}
\end{center}
\end{table*}

\begin{figure*}[h]
    \centering
    \vspace{-0.5 em}
     \includegraphics[width=.48\textwidth]{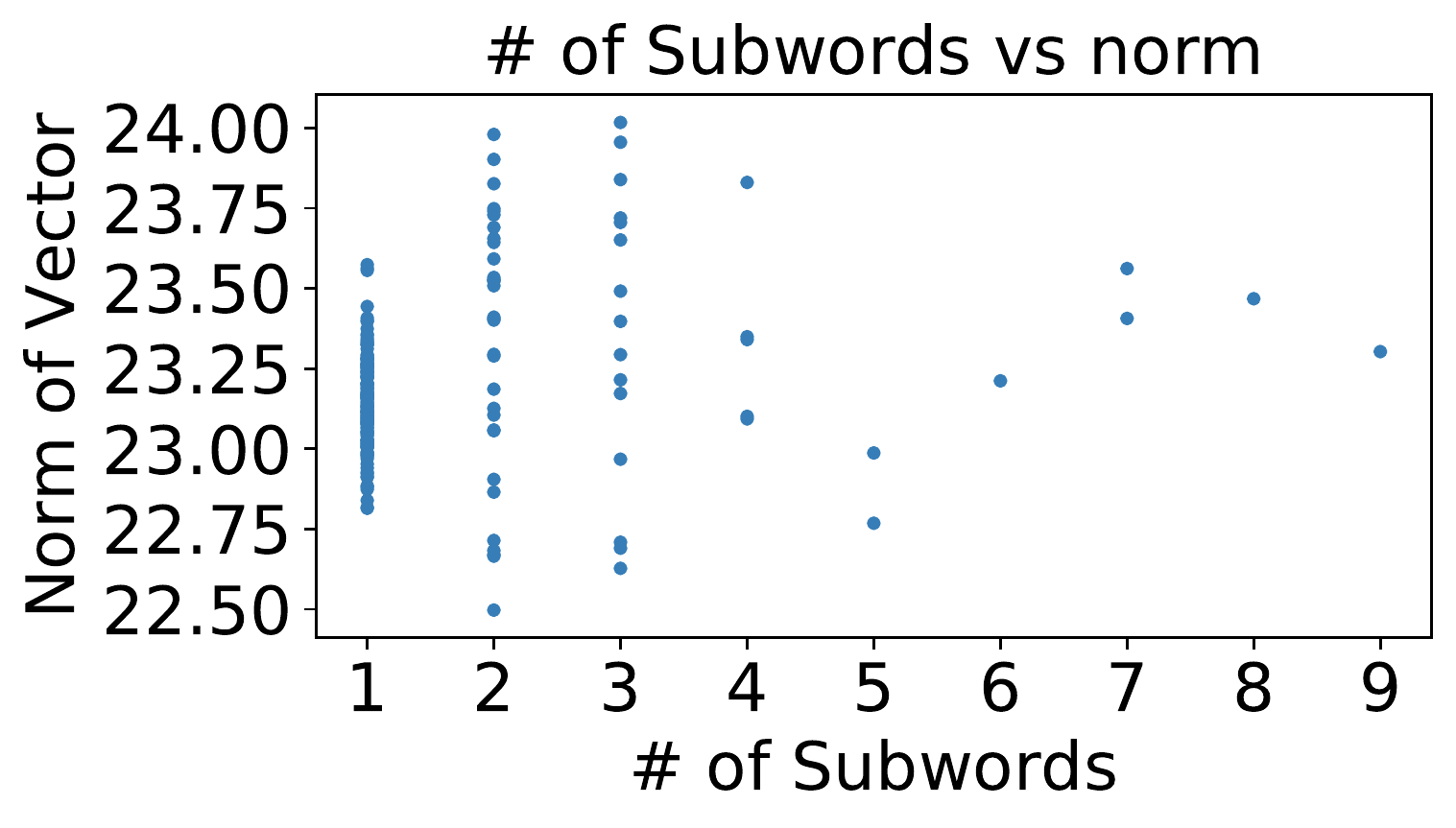}
     \vspace{-1 em}
     \caption{Average norm of embeddings in relation to the number of subpieces of the embedding. OOV words are represented by the first subword of the country name. Pearson's correlation between average norm and number of subwords, $R=0.22$.}
     \label{figure:norm_freq_first}
     \vspace{-0.5 em}
\end{figure*}

\begin{table*}[]
\begin{center}
\begin{tabular}{lccc}
\midrule
\textbf{Region} &
\textbf{\#} &
\textbf{In-Vocab}
 &\textbf{Population} \\
\midrule
Africa & 372 & 5\% & 23,296,502 \\
Americas & 594 & 7\% & 9,335,510 \\
Asia & 1,466 & 4\% & 34,046,146 \\
Europe & 372 & 25\% & 11,278,318 \\
N America & 49 & 50\% & 1,834,020 \\
Oceania & 20 & 75\% & 1,316,862 \\
\midrule
\end{tabular}
\caption{Average number of populous (>100,000) cities that are in BERT-base-cased's vocabulary.}
\label{table:cities}
\end{center}
\end{table*}

To illustrate the skew of BERT's vocabulary on a larger dataset, we repeat this experiment for names of cities across the world.\footnote{The city and population data was created by MaxMind, available from http://www.maxmind.com/.} Cities, similar to country names are also information-rich entities representing peoples and places. Filtering for populous (>100,000) single-word cities, 50\% of North American cities and 25\% of European cities are in-vocabulary compared to less than 6\% of city names from Asia, Africa, and Central and South America (Table \ref{table:cities}). Contrast Nigeria --- world’s fourth largest population of English speakers --- with the United Kingdom. Lagos is the only city in-vocabulary out of a possible 89 populous cities, compared to the United Kingdom where 53 out of its 64 populous cities are in vocabulary. 

\subsection{Appendix for section \ref{section:distinct}}
\begin{table*}
\begin{center}
\begin{tabular}{lclc}
\toprule
\textbf{Dep. Variable:}    &      Average Cosine Similarity      & \textbf{  R-squared:         } &     0.084   \\
\textbf{Model:}            &       OLS        & \textbf{  Adj. R-squared:    } &     0.077   \\
\textbf{Method:}           &  Least Squares   & \textbf{  F-statistic:       } &     12.13   \\
\textbf{Date:}             & Sat, 13 Nov 2021 & \textbf{  Prob (F-statistic):} &  0.000675   \\
\textbf{Time:}             &     11:05:47     & \textbf{  Log-Likelihood:    } &    330.05   \\
\textbf{No. Observations:} &         134      & \textbf{  AIC:               } &    -656.1   \\
\textbf{Df Residuals:}     &         132      & \textbf{  BIC:               } &    -650.3   \\
\textbf{Df Model:}         &           1      & \textbf{                     } &             \\
\bottomrule
\end{tabular}
\begin{tabular}{lcccccc}
                      & \textbf{coef} & \textbf{std err} & \textbf{t} & \textbf{P$> |$t$|$} & \textbf{[0.025} & \textbf{0.975]}  \\
\midrule
\textbf{Constant}     &       0.6973  &        0.019     &    37.114  &         0.000        &        0.660    &        0.735     \\
\textbf{freq\_logged} &      -0.0061  &        0.002     &    -3.482  &         0.001        &       -0.010    &       -0.003     \\
\bottomrule
\end{tabular}
\begin{tabular}{lclc}
\textbf{Omnibus:}       &  8.539 & \textbf{  Durbin-Watson:     } &    1.998  \\
\textbf{Prob(Omnibus):} &  0.014 & \textbf{  Jarque-Bera (JB):  } &    8.470  \\
\textbf{Skew:}          & -0.604 & \textbf{  Prob(JB):          } &   0.0145  \\
\textbf{Kurtosis:}      &  3.236 & \textbf{  Cond. No.          } &     113.  \\
\bottomrule
\end{tabular}
\caption{OLS Regression Results: Using training data frequency (logged) to predict the average cosine similarity between a country compared to all other countries (in-vocabulary countries only). Frequency explains 8\% of the variance. The more frequent the country, the lower average cosine similarity --- indicating its distinctness from all other countries.}
\label{table:freq_log_in_vocabulary}
\end{center}
\end{table*}

\begin{table*}
\begin{center}
\begin{tabular}{lclc}
\toprule
\textbf{Dep. Variable:}    &      Average Cosine Similarity      & \textbf{  R-squared:         } &     0.086   \\
\textbf{Model:}            &       OLS        & \textbf{  Adj. R-squared:    } &     0.079   \\
\textbf{Method:}           &  Least Squares   & \textbf{  F-statistic:       } &     12.47   \\
\textbf{Date:}             & Mon, 15 Nov 2021 & \textbf{  Prob (F-statistic):} &  0.000570   \\
\textbf{Time:}             &     16:46:42     & \textbf{  Log-Likelihood:    } &    440.48   \\
\textbf{No. Observations:} &         134      & \textbf{  AIC:               } &    -877.0   \\
\textbf{Df Residuals:}     &         132      & \textbf{  BIC:               } &    -871.2   \\
\textbf{Df Model:}         &           1      & \textbf{                     } &             \\
\bottomrule
\end{tabular}
\begin{tabular}{lcccccc}
                      & \textbf{coef} & \textbf{std err} & \textbf{t} & \textbf{P$> |$t$|$} & \textbf{[0.025} & \textbf{0.975]}  \\
\midrule
\textbf{Constant}     &       0.8458  &        0.008     &   102.629  &         0.000        &        0.830    &        0.862     \\
\textbf{freq\_logged} &      -0.0027  &        0.001     &    -3.531  &         0.001        &       -0.004    &       -0.001     \\
\bottomrule
\end{tabular}
\begin{tabular}{lclc}
\textbf{Omnibus:}       &  6.442 & \textbf{  Durbin-Watson:     } &    2.084  \\
\textbf{Prob(Omnibus):} &  0.040 & \textbf{  Jarque-Bera (JB):  } &    3.258  \\
\textbf{Skew:}          & -0.112 & \textbf{  Prob(JB):          } &    0.196  \\
\textbf{Kurtosis:}      &  2.270 & \textbf{  Cond. No.          } &     113.  \\
\bottomrule
\end{tabular}
\caption{OLS Regression Results: Using training data frequency (logged) to predict the average cosine similarity between a country compared to all other countries (in-vocabulary countries only) when names of countries appear in identical contexts. Frequency explains 9\% of the variance. The more frequent the country, the lower average cosine similarity --- indicating its distinctness from all other countries.}
\label{table:freq_log_in_vocabulary_artificial}
\end{center}
\end{table*}

\begin{table*}[]
\begin{center}
\begin{tabular}{l}
\toprule
\textbf{Sentence}   \\
\midrule
I am from COUNTRY.                                                             \\
I live in COUNTRY.                                                             \\
I hope this January I will get to travel to COUNTRY.                           \\
I am interesting in traveling to COUNTRY.                                      \\
My friend is from COUNTRY.                                                     \\
COUNTRY is well known for its history.                                         \\
COUNTRY has a diverse culture and a fascinating history.                       \\
COUNTRY has been involved in a number of historical events.                    \\
COUNTRY is developing its economic sector rapidly.                             \\
COUNTRY fought in a number of wars.                                            \\
Today my history teacher taught us about COUNTRY and its history.              \\
The geography of COUNTRY is fascinating.                                       \\
A number of scientists from COUNTRY have gained fame for their work.           \\
Living in COUNTRY definitely has its advantages and disadvantages.             \\
The government of COUNTRY is facing criticism.                                 \\
I never thought to visit COUNTRY until my neighbor told me about it.           \\
The news says that COUNTRY is going through some severe climate change.        \\
The athlete from COUNTRY has just won the Olympic medal.                       \\
The actress was born in COUNTRY and immigrated as a kid.                       \\
A number of fossils has been found in COUNTRY where scientists least expected.\\
\bottomrule
\end{tabular}
\caption{List of artificial sentences used in section \ref{section:distinct} to measure cosine similarity of country names in identical contexts.}
\label{table:artificial_sentences}
\end{center}
\end{table*}

\subsection{Appendix for section \ref{section:mlm}}

\begin{table*}[]
\begin{center}
\begin{tabular}{ccccc}
\toprule
\textbf{\begin{tabular}[c]{@{}c@{}}GDP\\ Quartile \end{tabular}} &
  \textbf{\begin{tabular}[c]{@{}c@{}}BERT\\ Base\end{tabular}} &
  \textbf{\begin{tabular}[c]{@{}c@{}}BERT\\ ML\end{tabular}} &
  \textbf{\begin{tabular}[c]{@{}c@{}}RoB-\\ ERTa\end{tabular}} &
  \textbf{Tuned} \\
  \midrule
1     & 26\%          & 15\% & 14\% & \textbf{28\%} \\
2    & 26\%          & 18\% & 22\% & \textbf{30\%} \\
3    & \textbf{32\%} & 22\% & 26\% & 30\%          \\
4 & \textbf{42\%} & 29\% & 39\% & 38\%         \\
\bottomrule
\end{tabular}
\caption{Performance on the MLM task as binned by gold label's GDP (by quartiles) from section \ref{section:mlm}. We see that performance is best on the high GDP countries and that our tuned model is able to perform better on the lower GDP countries which are in our set of interested countries (Table \ref{table:list_of_interested_countries}).}
\label{table:performance_by_model}
\end{center}
\end{table*}

\begin{table*}[]
\begin{center}
\begin{tabular}{lcc}
\toprule
\textbf{Country} & \textbf{GDP (millions)} & \textbf{Freq} \\
\midrule
India            & 2,891,582               & 505,003       \\
Iran             & 603,779                 & 160,699       \\
Poland           & 595,862                 & 197,231       \\
Egypt            & 317,359                 & 101,897       \\
Qatar            & 183,466                 & 22,766        \\
Angola           & 85,000                  & 19,839        \\
Myanmar          & 76,784                  & 23,840        \\
Uzbekistan       & 57,921                  & 13,244        \\
Serbia           & 51,475                  & 57,957        \\
Uganda           & 32,609                  & 33,860        \\
Cambodia         & 27,097                  & 22,095        \\
Iceland          & 24,188                  & 31,763        \\
Senegal          & 23,664                  & 14,672        \\
Syria            & 20,379                  & 53,987        \\
Jamaica          & 15,830                  & 40,257        \\
Madagascar       & 14,104                  & 23,512        \\
Bahamas          & 13,578                  & 13,882        \\
Guinea           & 12,354                  & 54,010        \\
Chad             & 11,271                  & 38,650        \\
Barbados         & 5,209                   & 13,829     \\
\bottomrule
\end{tabular}
\caption{List of twenty randomly selected countries from section \ref{section:mitigation}. Additional data for each of these countries was used to continue pre-training the model --- resulting in an increased in performance on this subset of countries.}
\label{table:list_of_interested_countries}
\end{center}
\end{table*}

\begin{table*}[]
\begin{center}
\begin{tabular}{lcc}
\toprule
\multicolumn{1}{l}{}          & \textbf{BERT-Base }         & \textbf{Tuned}           \\ \hline
\multicolumn{3}{l}{\textbf{Accuracy}}                                         \\ \hline
Interested Countries                   & 31.75\%            & 44.55\%         \\
Other Countries                     & 31.56\%            & 28.89\%         \\ \hline
\multicolumn{3}{l}{\textbf{\% of predictions}} \\\hline
Interested Countries                    & 17.62\%              & 44.44\%         \\
Other Countries                        & 82.37\%              & 55.56\%      \\
\bottomrule
\end{tabular}
\caption{Average accuracy MLM task of BERT-Base and our tuned model for our random set interested countries and all other countries. Average \% of predictions out of all countries predicted (non-countries words are not in denominator).}
\label{table:tuned_diff}
\end{center}
\vspace{-5mm}
\end{table*}

\begin{figure*}[]
    \centering
    \vspace{-0.5 em}
     \includegraphics[width=.48\textwidth]{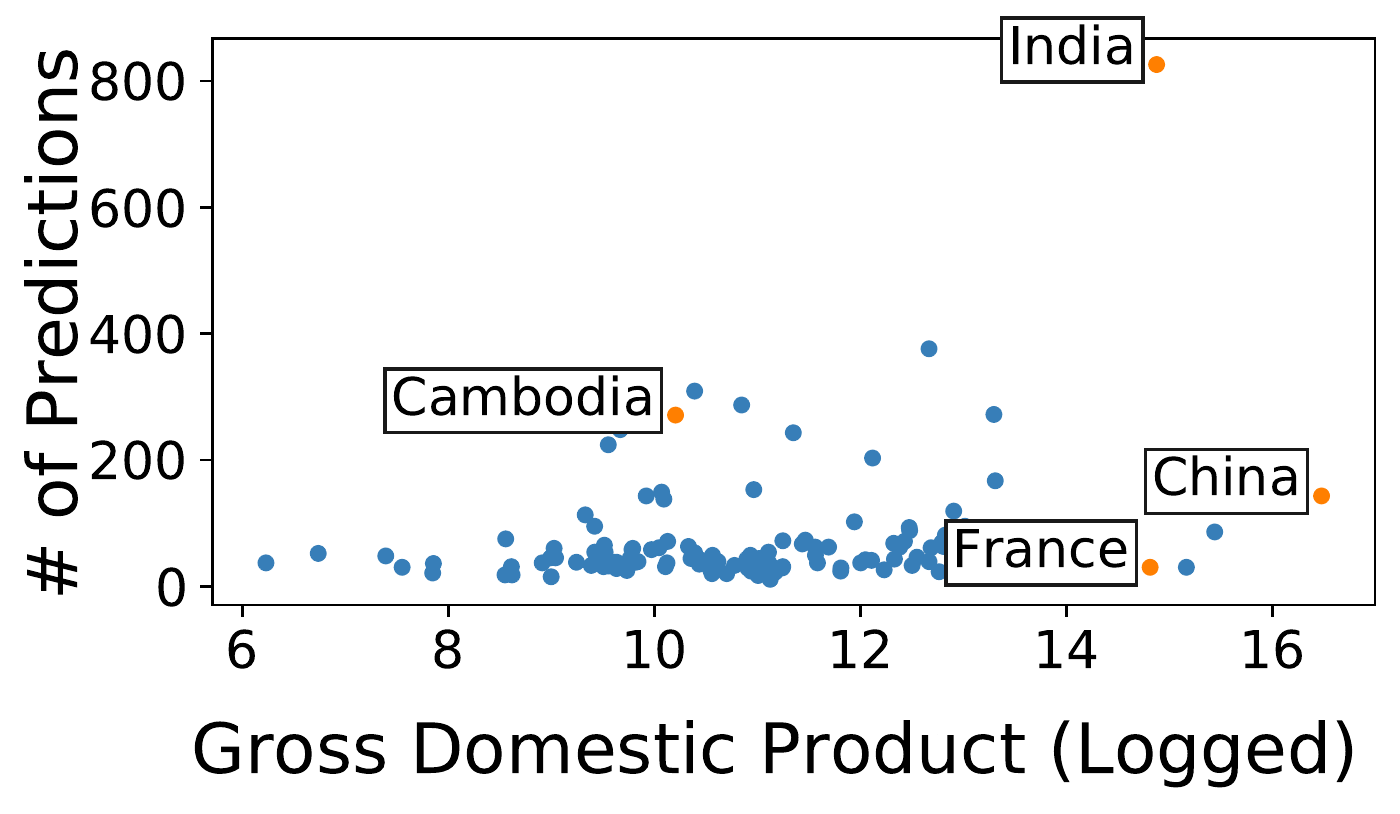}
     \vspace{-1 em}
     \caption{Number of times a country was predicted in the MLM task versus its GDP (logged) in our tuned model. Pearson's correlation between GDP (logged) and number of times a country is predicted for our tuned mode, $R=0.22$.}
     \label{figure:fig_tuned}
     \vspace{-0.5 em}
\end{figure*}

\end{document}